# Classification of Misinformation in New Articles using Natural Language Processing and a Recurrent Neural Network

**Written by Lydia Manikonda and Brendan Cunha**[1]

[1]Rensselaer Polytechnic Institute
110 8th Street
Troy, NY 12180
manikl@rpi.edu; cunhab@rpi.edu

## Abstract

This paper seeks to address the classification of misinformation in news articles using a Long Short Term Memory Recurrent Neural Network. Articles were taken from 2018; a year that was filled with reporters writing about President Donald Trump, Special Counsel Robert Mueller, the Fifa World Cup, and Russia. The model presented successfully classifies these articles with an accuracy score of 0.779944. We consider this to be successful because the model was trained on articles that included languages other than English as well as incomplete, or fragmented, articles.

## Introduction

Misinformation in news articles has been one of the main topics for discussion over the past few years. There have been several organizations that developed methods for assessing reliability and personal bias of news coverage. In today's day in age, it is unnatural to arbitrarily trust the news outlets that claim to be truly objective and unbiased because the term "bias" is relative. What one person perceives as bias, another may not; or rather what one person perceives as extremely biased and therefore untrustworthy, another may perceive as only slightly biased and therefore not too far to either side of the political scale to deem untrustworthy. Each person has a different threshold for judging just how much bias is "too much." This paper addresses this threshold in a logical way through a rating system based on ground truth scores aggregated by assessment sites. These assessment sites have crowd-sourced their data and will be referred to as "labels."

## Data Description

The dataset used consists of 713 thousand news articles collected between January, 2018 and November, 2018. These articles, stored in a SQLLite database, come from 194 different outlets including mainstream, hyper-partisan, and conspiracy sources. The ground truth aggregate scores come from 8 different assessment sites covering various scales of reliability, bias, transparency, and consumer trust. These ground truth labels come from the following sites stored in a CSV file:

- NewsGuard
- Pew Research Center
- Wikipedia
- OpenSources
- Media Bias/Fact Check (MBFC)
- AllSides
- BuzzFeed News
- Politifact

One of the first issues to address with these labels is the inconsistency of scales used. For example, some labels are scaled from 0-3 in terms of level of misinformation, others are scaled in a binary manner with 0 and 1, and some have 4 categorical values based on levels of media bias. So there is quite a bit of processing that needed to be done to normalize everything and transform the qualitative variables into quantitative variables.

Another issue is the discrepancy between trustworthy versus untrustworthy labels, and right political bias versus left political bias labels. During the pre-processing there was much thought on how to aggregate the political bias labels into trustworthy and untrustworthy classes; however, it was decided to omit these labels altogether and instead focus on the labels that strictly dealt with trustworthiness.

Aside from these two issues, the main problem encountered with this dataset was its size. It took quite a bit of time to clean and process the content of the articles, which will be discussed in detail later on in this paper. Also, it took multiple hours to produce word clouds, TFIDF cluster plots, trigram algorithms after removing stop words, and the Doc2Vec model. It took not as long, but still a decent amount of time, to run the Neural Network classification model to classify the articles as trustworthy or untrustworthy.

## Data Pre-Processing and Encoding

To pre-process the dataset, an in depth look at the CSV file of labels was accomplished. It was decided that the labels will be consolidated and encoded into columns with classes of either 0 for misinformation or 1 for trustworthy. Out of all the labels that focused on trustworthiness, they all included one column that consolidated all their aggregate scores into a binary value of 0 or 1. However, the aforementioned threshold amongst people is different for everyone; so to account for this we developed a rating for each article based on the encoded labels.

In the calculation, there were five labels used in the rating calculation: NewsGuard, Pew Research Center, Wikipedia, Media Bias, and Politifact. The rating was calculated based on the numbers of positive (1) and negative (0) scores for these labels, and then subtracting the negative score from the positive score. As a result, the ratings for the entire dataset ranged from -4 to 3, and the exact distribution of articles with these scores are as follows:

- -4: 23,279
- -3: 34,450
- -2: 72,055
- -1: 179,402
- +1: 104,043
- +2: 92,318
- +3: 46,822

These rating were then aggregated to a single binary column, where 1 represents a positive rating and is therefore trustworthy, and 0 represents a negative rating and is therefore untrustworthy. The distribution of this column can be seen below. It is clear that there are more negatively scored articles than positively, meaning that there are more articles classified as untrustworthy than trustworthy.

- 0 (untrustworthy): 309,186
- 1 (trustworthy): 243,183

After this, the dataset was split into two-thirds training and one-third testing. Then, both of these new partitioned datasets were cleaned with Natural Language Processing (NLP) on the column containing the articles. First, non-letters were removed and all text was converted to lower case. Then, punctuation was removed and the sentences were tokenized using nltk's "tokenizer" function. Lastly, stop words were removed from the text. The cleaning process omitted lemmentizing and stemming because of computing power; leaving these out of the cleaning process cut the running time in half, and made manipulation of the training and testing sets much easier and less time consuming. Cleaning the training data took around 16 minutes, and the testing data took around 9 minutes.

## NLP Data Exploration Analysis

The first task of data exploration analysis (DEA) on this dataset was to perform sentiment analysis. To do so, we first looked at polarity, which refers to identifying sentiment orientation in language. The goal was to separate objective expressions from subjective expressions, which would hopefully give us an idea of the underlying emotions in an article. Figure 1 shows a histogram of sentiment polarity for the entire dataset. The distribution's mean is right of 0, which means there is mostly positive sentiment underlying all of the articles.

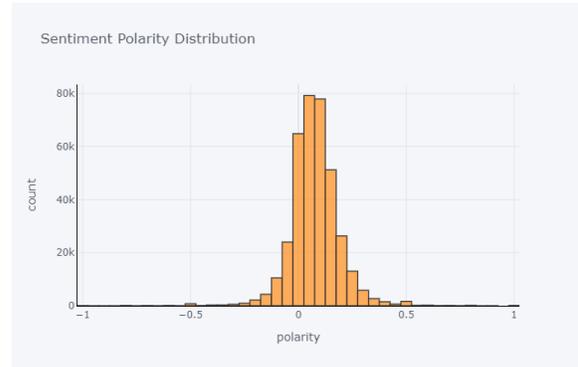

Figure 1: Sentiment polarity distribution for the entire cleaned dataset.

To give a sense of the success of the polarity algorithm, we produced 5 random articles with the highest positive sentiment polarity, which is 1 on a scale of -1 to 1. Some snippets from the articles are, "Declaring Tariffs are the greatest, President Donald Trump warned that..." and "Trump had his best 24 hours as President." Trump is the topic of these 2 articles as well as many others, which can be seen in the trigram plot in Figure 2. The top 5 trigrams from the entire dataset are "president donald trump," "new york times," "www thesun uk," "https www thesun," and "thesun uk sport." Some important events that happened in 2018 were the Robert Mueller investigation, the Fifa World Cup, and the investigation into the Russians meddling the 2016 presidential election.

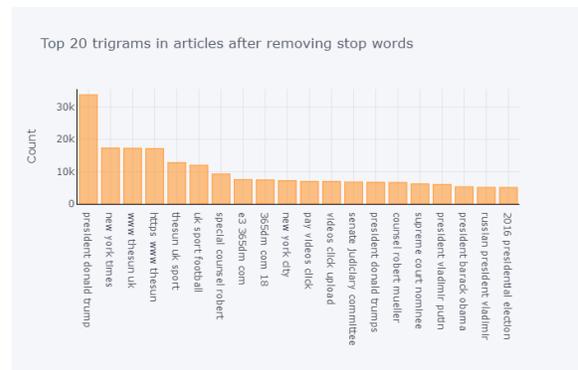

Figure 2: Top 20 trigrams for the entire cleaned dataset.

Additionally, we sought to obtain the top 20 part of speech tagging of the articles. Figure 3 shows a plot of this analysis.

Another form of DEA that was accomplished was topic modeling with Latent Semantic Analysis (LSA). We developed a matrix of topic counts per article where the rows represent the topics and the columns represent the articles. To reduce the number of rows, Singular Value Decomposition (SVD) was used, and then the articles were compared by taking the cosine of the angle between them. Listed below are the top 10 topics in the entire dataset, with the top 6 words from each. Notice how much of the topics coincide with the trigrams, but the topics give us a better idea of what exactly was being talked about in the articles during 2018. Based off of these topics, we produced a tSNE clustering plot to further dive into how the articles naturally clustered around these topics, which can be seen in figure 4.

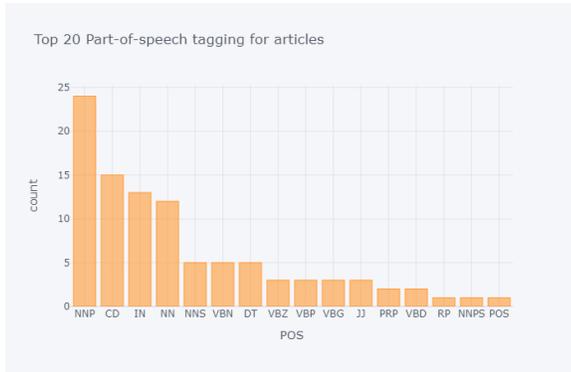

Figure 3: Top 20 parts of speech for the articles.

- Topic 1: com live cbsnews ftag www donald
- Topic 2: york new times city court jeff
- Topic 3: counsel robert mueller special https house
- Topic 4: president donald putin vladimir barack obama
- Topic 5: press white house secretary sarah york
- Topic 6: https www cbsnews york donald judiciary
- Topic 7: mr said trump abc court jeff
- Topic 8: committee court nominee supreme senate judiciary
- Topic 9: 2016 presidential election court jeff https
- Topic 10: abc attorney jeff told blasey christine

In addition, we obtained the top 10 topics amongst the different ratings described in the Data Pre-Processing and Encoding section. Listed below is the top topic from the -4, -2, +1, and +3 ratings.

- -4: general attorney jeff sessions deputy james
- -2: journalism read editorial accuracy standards truth
- +1: com 18 08 09 10 11
- +3: com live cbsnews ftag www donald

These types of analyses and figures are important to understanding the articles themselves and how to proceed with modeling. We decided build a Doc2Vec model using gensim and a recurrent neural network (RNN) using keras and sklearn.

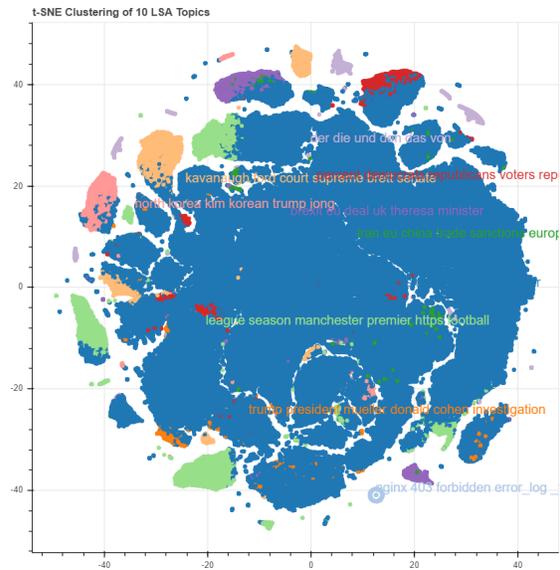

Figure 4: tSNE clustering using top 6 words.

## Modeling Part 1: Doc2Vec

Numeric representation of the articles was a challenging exercise, but was accomplished thanks to Doc2Vec. Understanding Doc2Vec is tied to understanding Word2Vec at a basic level. When building a model using words, labeling or encoding causes these words to lose their meaning. Word2Vec essentially produces a numeric representation for each word that will be able to capture relationships between them. For example, out of the words Troy, Albany, Rensselaer Polytechnic Institute, and University of Albany, numeric representations would relate Troy with Rensselaer Polytechnic Institue and Albany with University of Albany. Word2Vec is able to capture such representations through synonyms, antonyms, or analogies.

Instead of creating a numeric representation for the words in the articles, Doc2Vec creates a numeric representation of the articles themselves. The model was able to identify similar document adequately, and the output of the most similar articles using tags can be seen in the list below. The data was tagged in order to easily identify them when passing them through the model. So, think of the number in quotes as the tag and the decimal number as the similarity score. The training data was used to build the model, and then the results were inferred on the testing. Thus, the similarity score are the results from the testing data.

- ('217883', 0.9435918927192688)
- ('42113', 0.9166675806045532)
- ('307626', 0.9160533547401428)
- ('239757', 0.9154894351959229)
- ('26987', 0.9149177670478821)
- ('124239', 0.912755012512207)
- ('297007', 0.9125192165374756)

- ('254339', 0.912079930305481)
- ('73209', 0.9116336107254028)
- ('25726', 0.9113792777061462)

## Modeling Part 2: Long Short Term Memory (LSTM) Recurrent Neural Network

For the RNN modeling, we used Keras Embedding Layer and GloVe word embeddings to convert the text in the articles to numeric form. We randomly chose the number of layers, neurons, and all other parameters. We used the encoded aggregate column from the previous section that separates articles into binary classes of 0 and 1. The embedding layer converts the textual data into numeric data and is used as the first layer of the learning model. Firstly, we used the Tokenizer class from the Keras module to create a word-to-index dictionary. Each word ion the corpus is used as a key, while a unique index is used as the key value.

The GloVe embeddings were used to create a feature matrix. A dictionary was created to contain words as keys and their corresponding embedding list as values. The next object created was an embedding matrix where each row number corresponds to the index of the word in the corpus.

Since text is a sequence of words, a RNN was chosen instead of a CNN. We used an LSTM to solve sentiment classification issues in the dataset. We created an LSTM layer with 128 neurons. Figure 5 gives a summary of the model.

```
Model: "sequential_11"
_________________________________________________________________
Layer (type)                 Output Shape              Param #
=================================================================
embedding_11 (Embedding)     (None, 100, 100)          54239300
_________________________________________________________________
lstm_2 (LSTM)                (None, 128)               117248
_________________________________________________________________
dense_7 (Dense)              (None, 1)                 129
=================================================================
Total params: 54,356,677
Trainable params: 117,377
Non-trainable params: 54,239,300
_________________________________________________________________
None
```

Figure 5: RNN model summary.

The next step was to train the model and evaluate the performance on the test set. This produced a test score of 0.4730 and a test accuracy of 0.7556. The model output in Figure 6 shows the difference between the accuracy values for the training and test sets, which is very small and shows the model is not over fitting. Similarly, the same case is for the loss values, seen in Figure 7.

As a final test, we made a prediction on a single article to judge the success of the model. An article was randomly selected, and it ended up having a true classification of 1 meaning that it is trustworthy. To predict the trustworthiness of this article, we have to convert it into numeric form, which was done by using the Tokenizer that was created previously. Specifically, the text to sequence method converts the sentence into numerical form. Then, we padded our input sequences, which we also did for our corpus before. Lastly, we used the "predict" method of the RNN model and passed it our processed input article. This yielded a value of 0.779944. Since this value is greater than 0.5, we can say that the model produced a trustworthy classification, which is consistent with its true class of 1. The sigmoid function predicts values between 0 and 1 which why the number is a decimal. Inspiration for this model and test is credited to Usman Malik from StackAbuse.

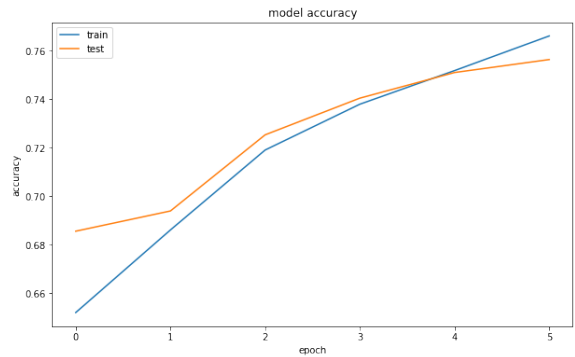

Figure 6: RNN model accuracy of train and test sets.

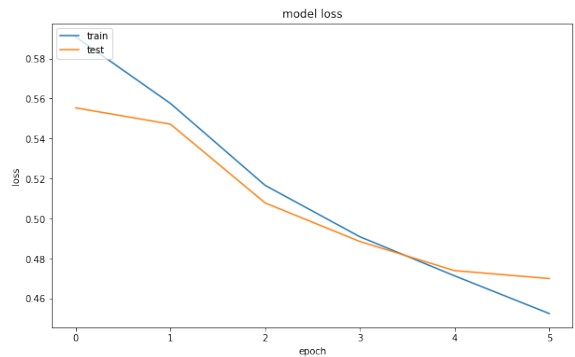

Figure 7: RNN model loss of train and test sets.

## Conclusion

To address misinformation in news articles, there are various methods for classification. In this paper, we sought to implement the most logical process to encode the labels into two columns: one being a binary classifier and one being a rating classifier that shows the level of trustworthiness. We build a Long Short Term Memory Recurrent Neural Network in order to classify the articles in the dataset, which yielded positive and accurate results. Moving forward, it would be nice to see these "trust" scores attached to news articles, so the general population can be made aware of the level of misinformation being presented in whatever news source they are reading.

## References

Malik, Usman "Python for NLP: Movie Sentiment Analysis using Deep Learning in Keras." https://stackabuse.com/python-for-nlp-movie-sentiment-analysis-using-deep-learning-in-keras/. 5/3/2020

Norregaard, Jeppe; Horne, Benjamin D.; Adali, Sibel. "NELA-GT-2018: A Large Multi-Labelled News Dataset for the Study of Misinformation in News Articles."

Shperber, Gidi. "A Gentle Introduction to Doc2Vec." https://medium.com/wisio/a-gentle-introduction-to-doc2vec-db3e8c0cce5e. 5/4/2020

Geron, Aurelien. "Hands-On Machine Learning with Scikit-Learn, Keras TensorFlow." O'Reilly. 9/2019